\documentclass[a4paper, 10pt, conference]{ieeeconf}      
\usepackage{FG2019}

\usepackage{amsfonts}

\FGfinalcopy 

\IEEEoverridecommandlockouts                              
\usepackage{graphicx} 

\def\FGPaperID{96} 

\title{\LARGE \bf
The Many Moods of Emotion}
\usepackage{cite}
\usepackage[utf8]{inputenc}
\begin{document}

\ifFGfinal
\thispagestyle{empty}
\pagestyle{empty}
\else
\pagestyle{plain}
\fi
\author{\parbox{16cm}{\centering
    {\large Valentin Vielzeuf$^1$ $^2$, Corentin Kervadec$^1$, Stéphane Pateux$^1$ and Frédéric Jurie$^2$}\\
    {\normalsize
    $^1$ Orange Labs, Rennes\\
    $^2$ Normandie Univ., UNICAEN, ENSICAEN, CNRS}}
}
\maketitle

\begin{abstract}
This paper presents a novel approach to the facial expression generation problem. 
Building upon the assumption of the psychological community that emotion is intrinsically continuous, we first design our own continuous emotion representation with a 3-dimensional latent space issued from a neural network trained on discrete emotion classification. The so-obtained representation can be used to annotate large in the wild datasets and later used to trained a Generative Adversarial Network. 

We first show that our model is able to map back to discrete emotion classes with a objectively and subjectively better quality of the images than usual discrete approaches. But also that we are able to pave the larger space of possible facial expressions, generating the many moods of emotion. 
Moreover, two axis in this space may be found to generate similar expression changes as in traditional continuous representations such as arousal-valence. 
Finally we show from visual interpretation, that the third remaining dimension is highly related to the well-known dominance dimension from psychology.
\end{abstract}

\section{Introduction}
\label{intro}
Affective computing is a topic of broad interest, finding applications in many fields such as health-care, marketing or human-machine interfaces. 
Therefore, a great effort has been put in the recognition of emotion across different contents. Indeed, several works propose to analyze facial expressions from images~\cite{ ng_deep_2015,acharya_covariance_2018}, from multimodal videos~\cite{ringeval2017avec,vielzeuf2017temporal,knyazev2017convolutional, dhall2018emotiw} or from multi-view videos~\cite{valstar2017fera,batista2017aumpnet}. Other works focus more on sentiment expressed within text~\cite{cambria2016affective,poria2016fusing, rosenthal2017semeval} or audio~\cite{schuller2013interspeech, schuller2018interspeech,huang2018speech}, finally building a very large and complete set of emotion recognition methods. 
Nevertheless, some recent works~\cite{vielzeuf2018occam} underline that the performance may begin to saturate on the emotion recognition task, because of the nature of the used datasets and of the subjective representation of emotion.
\begin{figure}[ht!]
    \centering
    \includegraphics[width=\linewidth]{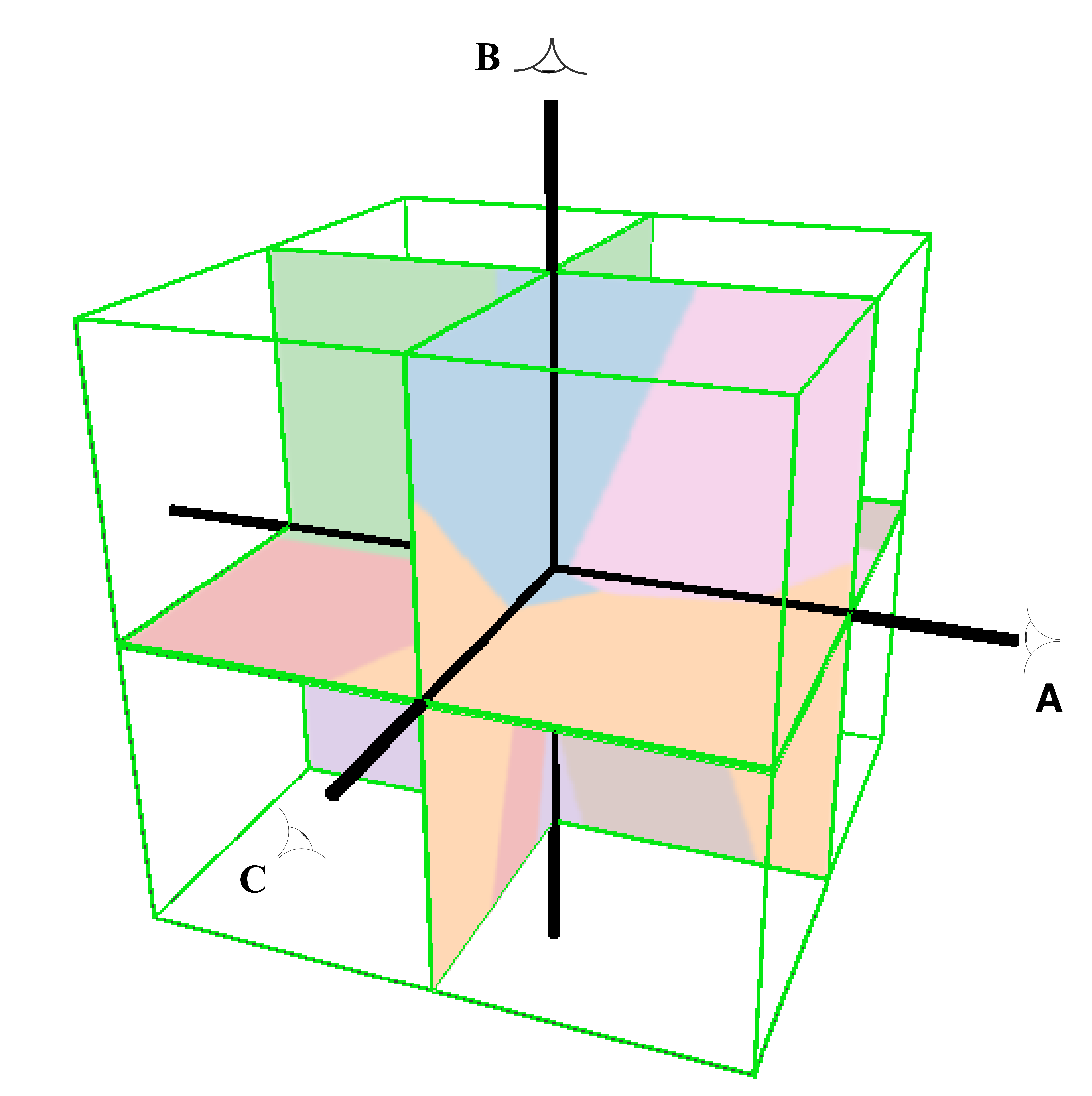}
    \caption{Illustration of our 3-d representation space of emotion. Better viewed in color.}
    \label{fig:manymoods}
\end{figure}

\begin{figure*}[ht]
    \centering
    \includegraphics[width=\linewidth]{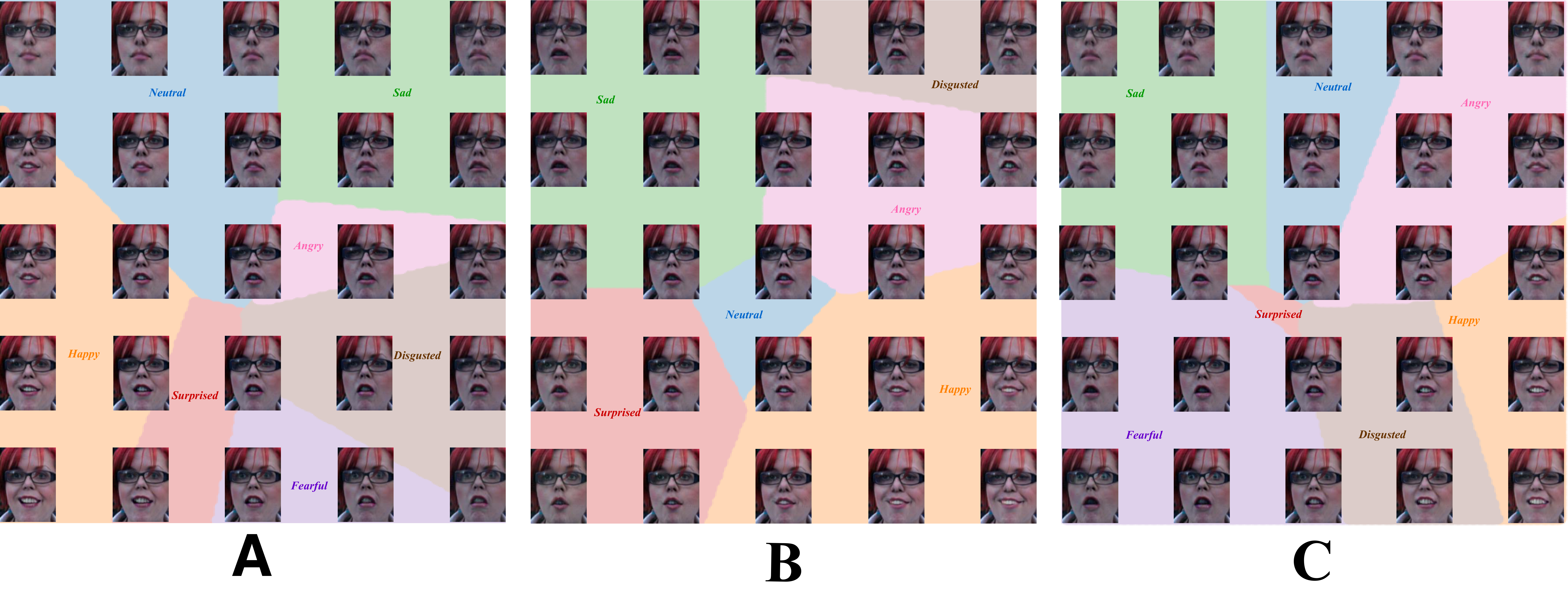}
    \caption{Illustration of our 3-d representation space of emotion. Each plane is colored with the associated discrete classes of emotions (see position of these planes in Figure~\ref{fig:manymoods}). Furthermore for each of these planes we illustrate some generated faces from the associated 3D representation coordinates. The generated sample faces within same color areas show the many possible moods inside a given emotion class. Note that expressions in these three planes are only a small part of the possible samples generated by our model. Better viewed in color.}
    \label{fig:manymoods2}
\end{figure*}

Thus, understanding and manipulating the emotion representation is of tremendous interest to progress towards a more complete affective computing ability. 
For that, the very definition of the facial expression of the emotion has to be taken into account. The literature comes up with three main definitions. First, Ekman \textit{et al.}~\cite{ekman_constants_1971} proposes discrete emotions, identifying six universal classes of emotion (e.g. "Happy", "Sad" or "Angry"). Later, the arousal-valence system was built by Russell, placing emotions in a 2-d continuous space. Finally, the Facial Action Coding Systems allows to objectively represent facial expression with Action Units (e.g. "raised eyebrow") and thus may be used to infer the emotion.

As the face is one of the main ways of expressing emotion, a branch of the affective computing community proposes to focus on the generation of facial expression, thus aiming at both a better interpretation of the visual consequences of an emotion representation and a straight-forward way to simulate emotion.
For generating artificial representation of emotions, first works from computer graphics community focus on the animation of the faces with model-based approaches~\cite{soladie2013invariant, weber2018unsupervised, kim2018deep}. More recently deep learning approaches and especially Generative Adversarial Networks~\cite{goodfellow2014generative,choi2017stargan,tulyakov2017mocogan, ding2017exprgan, qiao2018emotional} have been proposed, often borrowing ideas from the computer graphics community~\cite{song2017geometry,pumarola2018ganimation} but also aiming to learn these facial expressions from diverse datasets and representations. 
Nevertheless, these approaches are generally trained on small corpus with pronounced emotions~\cite{lyons1998japanese,lucey2010extended, zhao2011facial}, meaning that the space of possible generated expressions is limited.

Even if larger in the wild corpus using other emotion representations such as action units or arousal valence exist~\cite{li_reliable_2017,benitez-quiroz_emotionet:_2016,mollahosseini_affectnet:_2017}, they are only used by a few authors~\cite{pumarola2018ganimation}. Moreover, the annotation cost of such representation is really higher than for discrete emotion. 

Last but but not the least, the described approaches are often focusing on the quality of the generated faces. 
In this paper, we propose to add another concern, building a bridge between psychological interpretations of the emotion representation and what is visually observed. 
Our contribution is therefore to propose a 3-dimensional representation of emotion based on the latent space of a classification neural network. This model is trained on discrete emotion classification and thus required less annotation effort than for traditional continuous approaches. 
Using it to annotate a large in the wild corpus, we then learn a generative adversarial network on the so-obtained dataset. We show that not only the generation of faces is more robust than with other representations, but also that we could exhibit complementary directions within the 3D space representation that are in line with the common psychological definition of arousal, valence and dominance, enabling easy interpretations of the observed improvements.

\section{Related Works}
\paragraph{Emotion Representation}
Emotion representation is a well-explored topic in the psychological community, as mentioned in Section~\ref{intro}. Therefore, an easy way to build a taxonomy of the different emotion representations is to categorize them along two directions: semantic meaning and power of description.
The higher semantic meaning comes with discrete emotion~\cite{ekman_constants_1971}, each classes being associated with one word, but at the cost of loosing a lot of power of description, as behind one word many variations may be found. Proposing compound emotion~\cite{du2014compound} (\textit{e.g.} happily surprise) is a way to reach a more fine-grained representation while keeping a high-level semantic meaning. Nevertheless, even with a large vocabulary of words the whole space of emotion may not be completely described. Indeed, as shown by Russel~\textit{et al}, emotion is a continuum, thus requiring a continuous system to obtain a really fine-grained description. To keep a semantic meaning, several interpretable axis were proposed to build continuous spaces, such as arousal, valence or even dominance~\cite{mehrabian1996pleasure}.
At a much lower semantic level, but with a perfect depiction of the facial expression, the computer graphics community tends to propose a Facial Action Coding System, allowing to objectively represent facial expression with Action Units (\textit{e.g.} "raised eyebrow").

Using datasets annotated by representations with a great power of description allows to train more efficient model, but it implies a higher annotation cost. Our method is aiming at reaching a compromise between having a great power of description and a low-cost annotation process. 

\paragraph{Computer Graphics}
The face animation task has already been actively explored by the computer graphics community, some early works proposing 3D model-based approaches~\cite{blanz2003reanimating, yang2011expression}. More recently, Soladié~\textit{et al.}~\cite{soladie2013invariant} uses a 4-d emotion representation space to animate face and Active-Appearance Models features. In a more general fashion, Weber \textit{et al.}~\cite{weber2018unsupervised} proposes an unsupervised person-specific model which easily adapted to the targeted subject. 
Finally, hybrid approaches mixing deep learning and model-based method are also proposed.
Susskind \textit{et al.}~\cite{susskind2008generating} first proposes to train a deep belief network based on both action units and identity information to generate facial expression. More recently, another approach~\cite{song2017geometry} is using fiducial points to geometrically control the face animation while Tulyakov \textit{et al.}~\cite{tulyakov2017mocogan} is learning to directly generate sequences of images, based on a "content and motion" approach.
Quia \textit{et al.}~\cite{qiao2018emotional} use facial landmarks to improve the animation smoothness of a changing emotion. Kim \textit{et al.}~\cite{kim2018deep} enable to generate video face animation using another portrait video as an example.
These approaches are working on the very shape of the face and it therefore implies complex modifications of the model to adapt to "in the wild" conditions, where important illumination changes and occlusions are common. 

\paragraph{Generative Neural Networks}
To fulfill the previous requirement of robustness towards real "in the wild" conditions, an interesting path of research for image synthesis using neural networks is Generative Adversarial Networks(GAN)~\cite{goodfellow2014generative} and Variational AutoEncoders(VAE)~\cite{kingma2013auto}. Focusing on GANs, many extensions exist, such as Conditional GAN~\cite{mirza2014conditional} where a condition variable allows to control the generation or more recently StarGAN, where Choi \textit{et. al}~\cite{choi2017stargan} propose a multi-domain approach, learning both facial attribute transfer and facial expression generation. Interestingly, the targeted facial expression is fed with the input face to modify, allowing an end-to-end approach.
Extending the previous works, Ding~\textit{et al}~\cite{ding2017exprgan} propose a new GAN framework enabling to learn intensity of an emotion by a specific encoding of the emotion label. The covered domain of possible facial expression is then larger than for classical discrete approaches, each class containing many variations along an intensity criteria. Nevertheless, this approach does not allow to generate all the possible facial expressions such as compound emotions. Pumarola~\textit{et al.}~\cite{pumarola2018ganimation} propose a more general approach, coupling GAN and Action Units to continuously generate facial expressions from a large dataset. This implies a lot of labeling work, as action units are costly to annotate. Moreover the constructed space has a high dimension (15 action units) leading to non direct analysis of the organization of the generated faces.

\section{Methods}
\label{methods}
This section describes our proposed approach. First we present how to exhibit a continuous representation of the emotions. The aim of this continuous representation being to enable compound emotions but also other possible variations within a given emotion. Then from this representation we adapt a GAN to allow for continuous emotion editing.

\subsection{Facial Expression Representation} 
As argued by~\cite{russell_circumplex_1980} continuous annotations may be a more subtle and accurate emotion representation than discrete basic emotion. The idea behind is to continuously quantify several features of a facial expression, such as intensity (arousal) or pleasure (valence). A mapping to the discrete representation of emotion may be done~\cite{russell_circumplex_1980,mollahosseini_affectnet:_2017}, enabling to take benefits from both discrete and continuous annotations. Nevertheless, as underlined by psychological study~\cite{mehrabian1996pleasure}, two dimensions might not be sufficient to represent the whole variations of emotions. Moreover, annotation cost is high compared to discrete emotions. But a recent approach~\cite{kervadec2018cake} shows that a compact latent space issued from intermediate hidden layers of a convolutional neural network trained on discrete emotion classification may lead to an arousal-valence like topology.

\begin{figure}[ht]
    \centering
    \includegraphics[width=\linewidth]{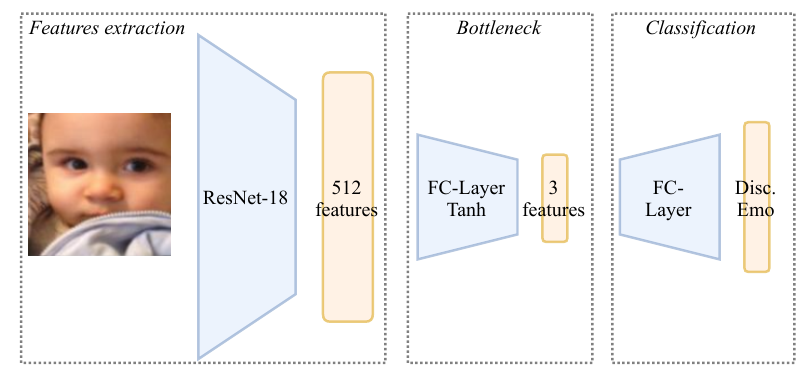}
    \caption{Generating a 3-d compact representation, following a similar approach to~\cite{kervadec2018cake}}.
    \label{fig:cake}
\end{figure}

We therefore propose to use a latent space of a convolutional neural network. In order to assess and understand the benefit of using additional dimensions in emotion representation~\cite{soladie2013invariant}, we propose to define a 3-d representation from the latent space. For that, we train a modified ResNet-18~\cite{he_deep_2015} to classify discrete emotions, as in Figure~\ref{fig:cake}. The modification consists in adding a bottleneck fully-connected before classification, forcing the classifier to use only three features to predict the discrete emotion. An hyperbolic tangent activation is applied on these three features to ensure that the 3-d representation range is kept between -1 and 1. 

From an initial corpus annotated only in discrete emotions, we train such an aforementioned network. This network is then used to provide continuous annotations to any provided dataset. 
In our case we only consider the dataset initially annotated in discrete emotion, but we could also have use any additional faces without any need of annotation.

We may observe in the upper left of the Figure~\ref{fig:manymoods} how the discrete emotions are associated to the 3-d representation in our representation space and especially on three planes cutting it. The many variations observed in these planes are also reported in the Figure~\ref{fig:manymoods2} as examples of the diversity of possible generated expressions.

\subsection{Facial Expression Generation}
To achieve facial expression generation, we choose to build a Generative Adversarial Neural Network (GAN). Among the various GAN approaches, we retain the StarGAN~\cite{choi2017stargan} architecture, which allows to take both the face and the targeted emotion as input of the generator and has already prove to be efficient on discrete emotion generation. As a reminder, the model is composed of a discriminator D and a generator G. Thus, as in the original approach, we use a loss composed with different terms. Nevertheless, we need to adapt them to the continuous labeling case. We therefore rewrite the following terms.

The \textit{adversarial loss} aims at making the generated fake expressions not distinguishable from real facial expressions. It may be written following:
\begin{equation}
    L_{adv} =  \mathbb{E}_x[log D(x)] + \mathbb{E}_{x,r}[1-log D((G(x,r)))]
    \label{adv}
\end{equation}

where $x$ is the input image and $r$ is the 3-dimensional facial expression representation. The generator and the discriminator respectively aim at minimizing and maximizing the term.

The usual classification loss, which we denote here as a \textit{regression loss}, is itself composed of two terms. The first term, namely $L^{real}_{reg}$, forces D to correctly regress the ground truth associated to the real image. While the second term, namely $L^{fake}_{reg}$, forces G to generate facial expressions with a representation close to the target. More formally, we use a Mean Squared Error term:
\begin{equation}
    L^{real}_{reg} =  \mathbb{E}_{x,r}[D(x)-r]^{2}
\end{equation}
\begin{equation}
 L^{fake}_{reg} =  \mathbb{E}_{x,r}[D(G(x,r))-r]^{2}
 \end{equation}

The \textit{reconstruction loss}, namely $L_{rec}$ ensures that the generated faces preserved contents not relative to the expression. It is defined as:
\begin{equation}
    L_{rec} = \mathbb{E}_{x,r_1,r_2}[||x-G(G(x,r_2),r_1)||_1]
\end{equation}
where $r_1$ is the original facial expression representation and $r_2$ is the representation of the facial expression the generator has to generate.

Finally, we can then write the generator and discriminator losses as:
\begin{equation}
    L_D = -L_{adv} + \lambda_{reg} L^{real}_{reg}
\end{equation}
\begin{equation}
    L_G = L_{adv} + \lambda_{reg} L^{fake}_{reg}+\lambda_{rec}L_{rec}
\end{equation}

For efficient implementation, the Equation~\ref{adv} can be reformulated with a Wasserstein GAN objective with gradient penalty, as done in the original paper~\cite{choi2017stargan}.

\subsection{Implementation Details}
To build the training set, we use the recent AffectNet dataset~\cite{mollahosseini_affectnet:_2017} which provides both discrete emotions and arousal valence annotations. We sanitize the dataset by discarding samples where there is no face or where the provided annotation does not exist. The resulting practical dataset then consists of 297000 annotated faces recorded in the wild. We preprocess them, using a face detector and a landmark aligner. To fit our convolutional neural networks requirements, faces are then resized to 256x256x3. During training time, we also apply augmentation transformations (scale jittering, rotation and flip).

We train (on the training set) our modified ResNet-18 and thus obtain our 3-d representation for each face.
Then, we train three GANs (discreteGAN, avGAN, and ours), one for each annotation type (resp. discrete, arousal-valence, and ours). We use a batch size of 16, a learning rate of 1e-4 with exponential decay factor of 0.996. The architectures of both generator and discriminator are similar to the one described in~\cite{choi2017stargan}. In the case of continuous annotations, the regression loss ponderation $\lambda_{reg}$ is changed to 3 instead of 1 because of the scale difference with a classification cross-entropy loss. The other ponderations are the same as in~\cite{choi2017stargan}. The parameters are optimized with the Adam method during 300000 iterations. 

To evaluate the generation task, we use the test set faces of AffectNet and generate faces of size 128x128x3.

\section{Results}
This section evaluates the benefits of the proposed continuous approach. We first detail our experimental protocol and then present our results. We study the ability of our approach to be applied to discrete emotion generation, then we enlighten the relations between our learned representation and the arousal valence representation. Finally, we build a bridge towards psychological interpretations, finding back a third dimension visually similar to the dominance~\cite{mehrabian1996pleasure}.

\begin{figure}[ht]
    \centering
    \includegraphics[width=\linewidth]{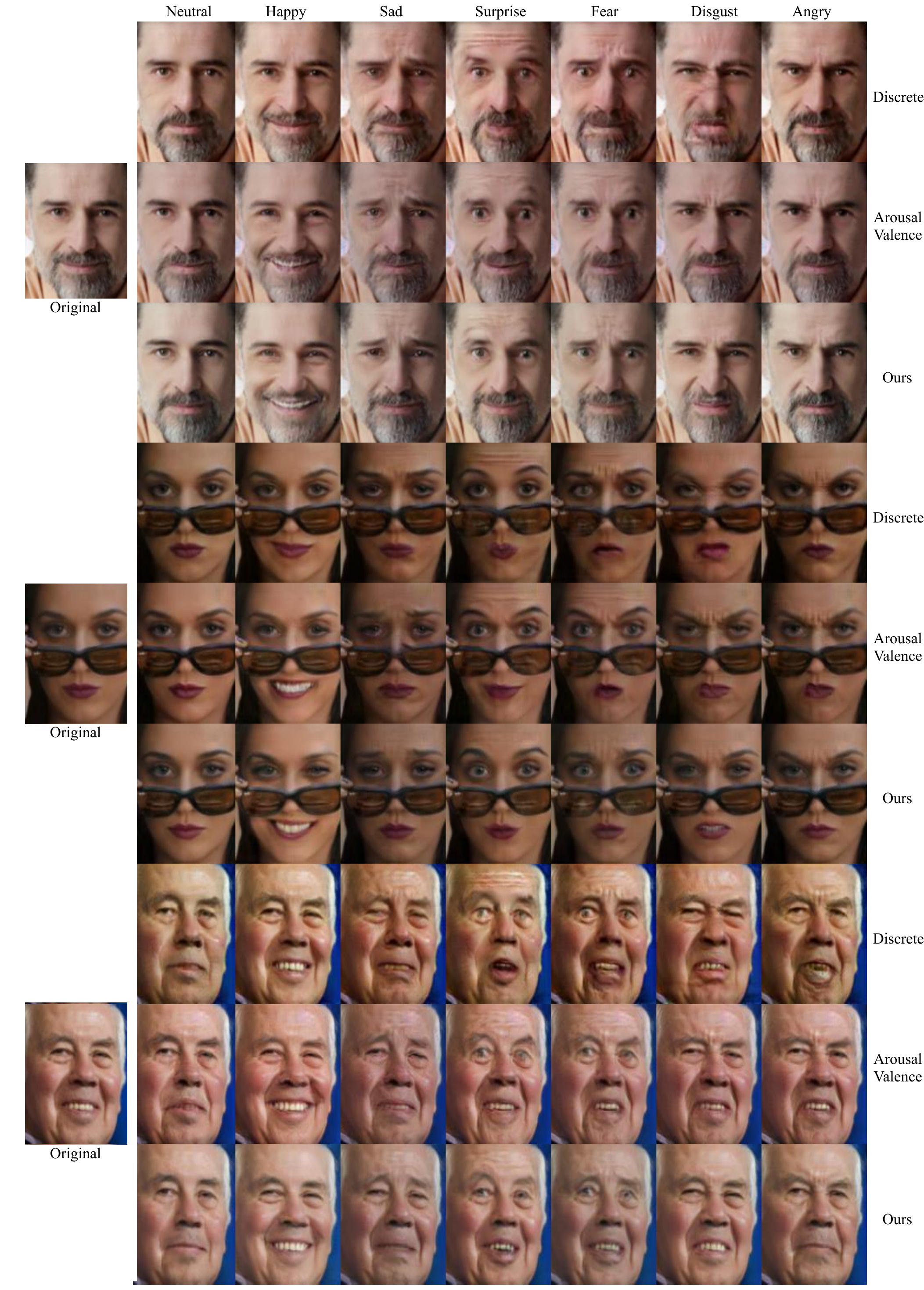}
    \caption{The 7 emotion classes generated with the three different approaches: discreteGAN (first row), avGAN (second row) and ours (third row). The three used examples are randomly extracted from the test set, to ensure a fair comparison between the approaches.}
    \label{fig:emotion}
\end{figure}
\subsection{Discrete Results}
\begin{figure*}[ht]
    \centering
    \includegraphics[width=\linewidth]{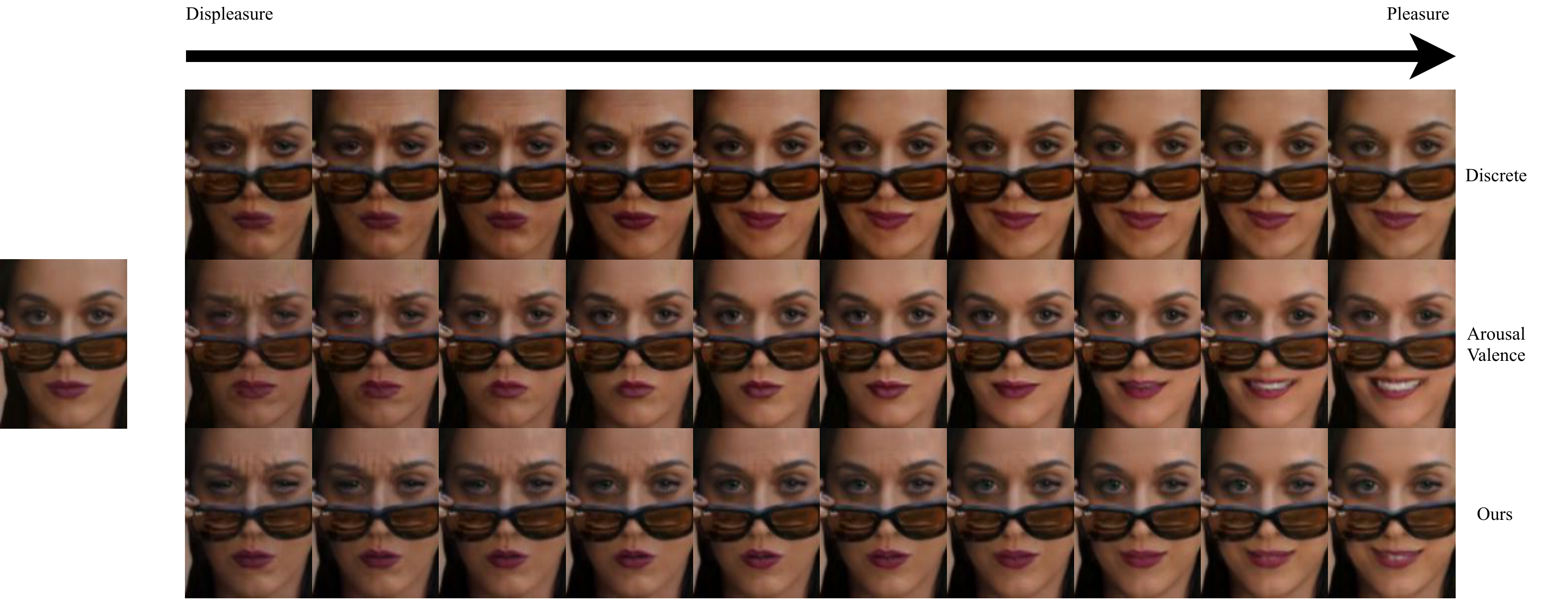}
    \caption{Generation of expression along the valence axis, from displeasure to pleasure. First row is discreteGAN using the same approach as in~\cite{ding2017exprgan}, second row is avGAN and third row is ours, using a linear regression to find a similar axis to valence. The input image is on the left. }
    \label{fig:valence}
\end{figure*}
We propose to evaluate the impact of the different representations on the quality of the generated expressions. Therefore, the same previously described GAN architecture is used -- we are not comparing different architectures of GAN -- and we only vary on the use of emotion representation. 
For that, we first generate the seven discrete emotions, using the discreteGAN as baseline.
Then, we choose to compute the coordinates of the centroid in the continuous representation spaces for each emotion class. We  thus simply generate the expression associated to these centroids and we label them with the emotion classes. More formally,
\begin{equation}
    C_{continuous}^{i}=\sum_{k \in C_{discrete}^{i}} \frac{r_{k}}{\#C_{discrete}^{i}}
\end{equation}
where $C_{continuous}^{i}$ is the coordinates of the centroid of the class $i$, $C_{discrete}^{i}$ is the set of all elements of the class $i$, and $r_k$ is our continuous representation of the sample $k$.
As we can observe in the column of Figure~\ref{fig:emotion}, the generated emotion classes are relatively similar for all GANs. Nevertheless, we may denote some interesting differences. 

For the happy class (second column), the continuous models tend to add teeth to the faces, with a better success for our approach than for avGAN (some artifacts are visible for the second face in the case of avGAN). 

In the case of the disgust class, we note the presence of artifacts in the faces generated by the discreteGAN, while avGAN tends to generate relatively similar expression for both disgusted and angry faces. The artifacts may be explained by the small number of disgust occurrences in the training dataset (less than 2\%), meaning that the discreteGAN did not see many examples of this class. Furthermore disgust and anger classes are relatively closed in arousal valence space, leading to very similar values for their centroids and thus very similar generated expressions. Our GAN, using a third dimension as shown in Figure~\ref{fig:manymoods}, allows to improve the separation of these two classes and thus leads to a real difference between the generated faces. 

The neutral class is also interesting, especially for the third face, where we can note the ability of our GAN to improve the control of the mouth closing. 
We also can note that the intensity of the expressions is higher in the discreteGAN faces. It may be explained by the fact that we choose the centroids for continuous representation, which thus are not extreme samples of these classes. 
\begin{table}[ht]
\centering
\begin{tabular}{l|llll|l}
\hline
         & \multicolumn{4}{l|}{RMSE on mean color}                      & $L_{rec}$          \\ \cline{1-5}
GAN      & Red          & Green        & Blue         & All         &               \\ \hline
Discrete & 4.5          & 6.3          & 10.2         & 7            & 0.22          \\
AV       & 6.4          & 7.8          & 5.7          & 6.7          & 0.14          \\
Ours     & \textbf{3.7} & \textbf{3.1} & \textbf{3.2} & \textbf{3.4} & \textbf{0.12} \\ \hline
\end{tabular}
\caption{Evaluation of the reconstruction quality and color conservation (in RMSE) of the different approaches on the test set. Lower is better.}
\label{rmse}
\end{table}

Finally, we may think from the third face that the discreteGAN is changing the mean color of the faces it generates. Another good qualitative example can be seen on Figure~\ref{fig:arousal}. 
To be more objective and assess if this visual observation is true, we compute the mean color value of the original image and and the mean values of the generated images of the seven emotion classes. We measure the root mean square error between the original mean color and the mean of the mean colors of the seven generated images on the whole test set (5000 faces) for each GAN in Table~\ref{rmse}. We observe that the error is really lower in our case and that there is a clear difference on the blue channel between discreteGAN and continuous GANs. These observations are in line with reconstruction losses (cycle-consistency loss with L1 norm, as in the original StarGAN~\cite{choi2017stargan}) obtained by the different GANs. 
Nevertheless, when generating a new expression, the observed color change may be explained by a bias learned by the model. For instance, negative emotions are often associated to a darker context and some expressions may imply a color modification, such as teeth showing during a smile. 
Finally, the last column of Table~\ref{rmse} reports the $L_{rec}$ evaluated on the test set for the different GANs and objectively show that the face is better preserved in the case of our GAN.

\subsection{Continuous Results}
\begin{figure}[ht]
    \includegraphics[width=\linewidth]{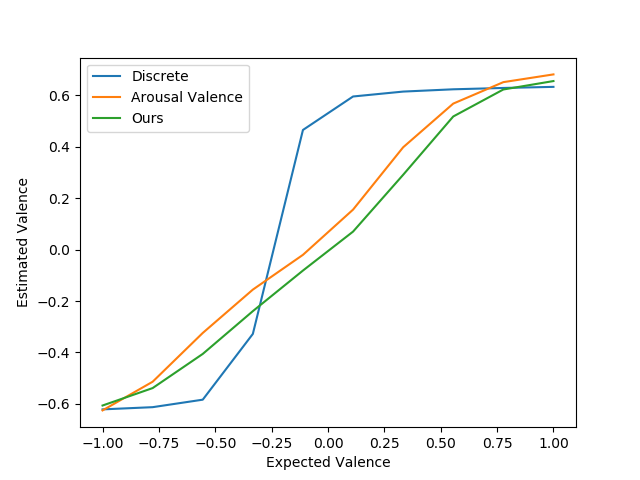}
    \includegraphics[width=\linewidth]{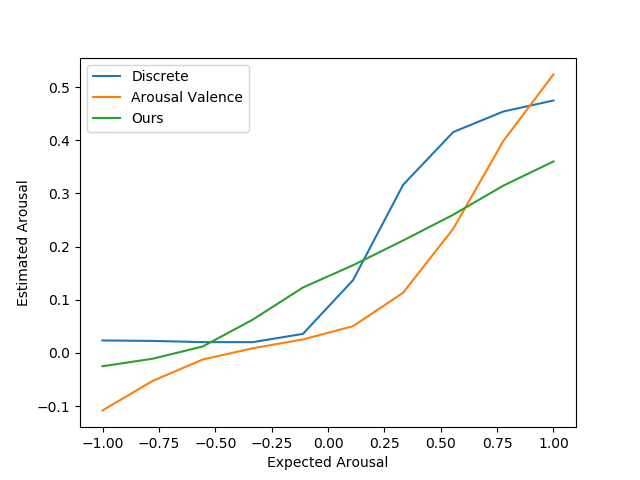}
    
    \caption{Plots of respectively the estimated valence (up) and arousal (down) of the generated faces as a function of the targeted valence (up) and arousal (down) by the model for the different approaches. This plots are average plots on the whole test set. Better viewed in colors.}
     \label{fig:curves}
\end{figure}

We are now focusing on the ability of the different methods to generate transitions between different expressions. To be able to compare the different methods, we choose to evaluate the transitions on arousal and valence axes, which are easy to interpret and often used in the psychological community~\cite{russell_circumplex_1980}.
For avGAN, the transitions is therefore straight-forward, we only need to browse through the different values of one dimension.
To generate continuous transition with the discreteGAN, we encode the emotion in a one hot vector and therefore create variations between two one hot vectors, as previously proposed by~\cite{ding2017exprgan} who are using this concept to vary "in intensity" between neutral class and another emotion. For the valence axis, we choose a transition between sadness (valence equals -1 and arousal close to 0)  and happiness (valence equals 1 and arousal close to 0), while for the arousal axis, the transition is between neutral (arousal and valence to 0) and surprise (valence close to 0 and arousal equals 1)\footnote{so it means there that we are not ranging in the complete [-1,1] interval but in [0,1])}.
For our GAN, we find the 3-d coordinates of the transition axis by applying a linear regression between our representation and arousal valence.

From Figure~\ref{fig:valence}, we first can observe that all the generated faces respect the valence axis, expressing displeasure at the extreme left and pleasure at the extreme right. We also can note that the expression from one GAN to another are not exactly similar, which may also be explained by the fact that the chosen axis for discreteGAN and for our GAN are not perfectly fitting the valence axis. 

Another important point is about the smoothness of the transition. Looking at the first line of Figure~\ref{fig:valence}, we observe four very similar expressions of displeasure, followed by one or two expressions mixing both displeasure and pleasure and finally five close expressions of pleasure. In the contrary, for both avGAN and ours, the transitions is smoother, the expression being modified at each face. So this would mean that the discreteGAN is not able to uniformly fit the axis of valence and tends to generate less variety in the expressions. 

\begin{figure*}[ht]
    \centering
    \includegraphics[width=\linewidth]{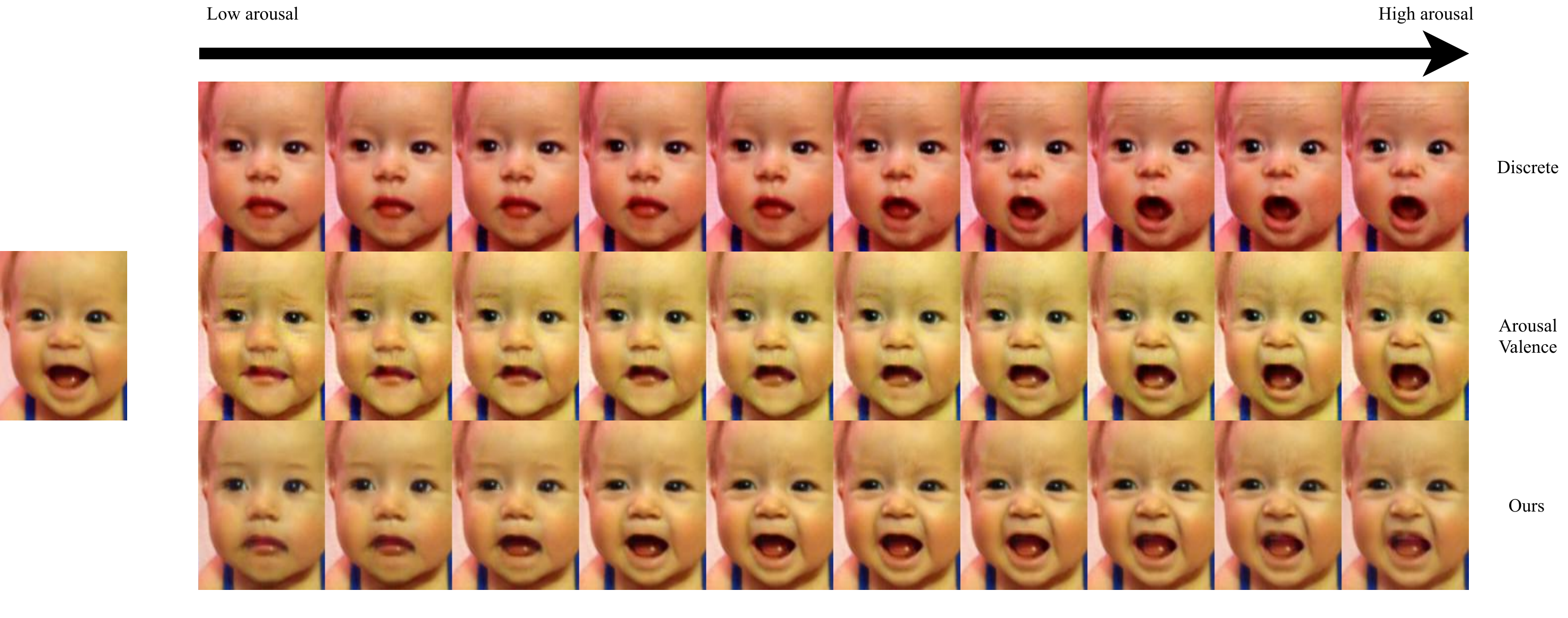}
    \caption{Generation of expression along the arousal axis, from displeasure to pleasure. First row is discreteGAN using the same approach as in~\cite{ding2017exprgan}, second row is avGAN and third row is ours, using a linear regression to find a similar axis to arousal. The input image is on the left. Better viewed in color.}
    \label{fig:arousal}
\end{figure*}

To verify this hypothesis, we propose to use a more objective process. First, we train a ResNet-18 to predict arousal and valence of the faces of AffectNet~\cite{mollahosseini_affectnet:_2017}. Second, we use this model to estimate the valence of the generated faces. Thus, we can plot the estimated valence in function of their targeted valence. We report the mean plots on the whole test set in Figure~\ref{fig:curves}. The avGAN's curve (in orange) should be the identity if both the arousal valence estimator and the avGAN were perfect. It is not the case, but we nevertheless can check that the allure of the curve is coherent with this idea. The curve of our GAN (in green) has a similar allure, validating the smoothness of the expression transition observed on Figure~\ref{fig:valence}. Finally, the discreteGAN's curve (in blue) has an allure which is closer to a step function than to the identity. It is also in line with what has been visually observed and highlight the fact that the discreteGAN is not suited to build a uniformly sampled space of representation.

From Figure~\ref{fig:arousal}, we note that all the GANs are able to generate a transition from a not excited face to a really aroused expression. As observed for the valence axis, the expressions are not totally similar from one GAN to another, as they are not generating expressions exactly on the same axis of arousal.
We can also observe again that the discreteGAN transitions is not very smooth in arousal. To assess this idea, we apply the same process used for valence and plot in Figure~\ref{fig:curves} the estimated arousal as a function of the targeted arousal. This plot first enlightens that the discreteGAN (blue), as for the valence, is not fitting an identity function. We also need to note the range of the estimated axis: the arousal values should be between -1 and 1 and are between -0.1 and 0.6 in the best case (avGAN). This may be explained by the fact that there are almost no sample with a negative arousal in the training set, leading both the estimator and the GANs to be inefficient on negative values.

\subsection{Interpreting the Third Dimension}
\begin{figure*}[ht]
    \centering
    \includegraphics[width=\linewidth]{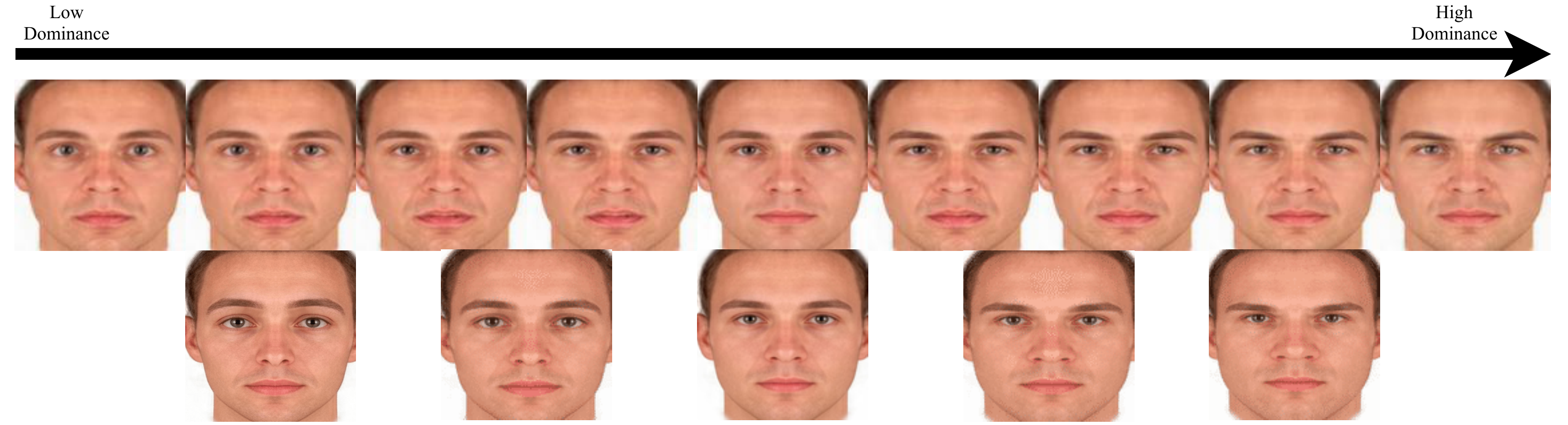}
    \caption{Illustration of the third dimension found from our representation and used to generate expressions in the first row. Second row is a manual work~\cite{van2015many} illustrating what dominance is.}
    \label{fig:dominance}
\end{figure*}
Even if the psychologists' community proposes arousal valence for emotion representation, several works show the limitations of using only two dimensions. Therefore, supplementary dimensions have been proposed and one is especially used. It is called the dominance~\cite{mehrabian1996pleasure} and may be seen as a measure of self-confidence.
In the previous sections, we show that our representation allows to map back to both discrete emotions and arousal valence. As already observed in Figure~\ref{fig:emotion}, it is difficult to distinguish disgust from anger with arousal valence representation, which is not the case with our 3-d representation. The third dimension may therefore brings interesting information. To dig into this idea, we propose to take the two directions of arousal and valence in our 3-d representation that we previously regress. From the two so-obtained vectors, we compute a third orthogonal vector (by vector product) and generate the expressions along this new axis.

The Figure~\ref{fig:dominance} displays in the first row the generated expression on the dominance axis. As our corpus is not annotated with the dominance value, we propose to compare our generated expressions to manually generated expressions proposed by Allen Grabo~\cite{van2015many}~\footnote{https://allengrabo.myportfolio.com/shifting-personality} (second line).
Even if the reconstruction is not perfect, we visually can see the same evolution in the facial expression, the self-confidence growing from left to right. This a first hint to show that our representation contains the dominance information, which has been learned from discrete labels.

\section{Conclusion}
We propose a solution to the facial expression generation based on a specific emotion representation containing 3 dimensions. 
This continuous representation is obtained from the constrained latent space of a neural network trained on the discrete emotion classification task. The so-obtained neural network can be used to annotate every face corpus such as in the wild datasets, allowing to learn from continuous representation. 
We therefore train a Generative Adversarial Network with these annotations. For that, we modify a well-known StarGAN architecture to fit the requirement of a regression approach. 
The obtained generated faces are compared to the same architecture trained in the same conditions but with other representations. We show qualitatively and quantitatively that not only our generated faces have a better reconstruction quality than a GAN trained on discrete emotion, but also that we are able to uniformly fit arousal and valence axis, as a GAN that would have been trained on real arousal valence labels.
Moreover, we exhibit a third dimension close to the concept of dominance, building a bridge with psychological interpretations of emotion.

\addtolength{\textheight}{-1cm}   

\bibliographystyle{ieee.bst}
\bibliography{biblio}

\end{document}